\documentclass[letterpaper, 10 pt, conference]{ieeeconf}
\usepackage[utf8]{inputenc}
\usepackage[T1]{fontenc}

\IEEEoverridecommandlockouts
\usepackage{graphicx}
\usepackage{times}
\usepackage{amsmath}
\usepackage{amssymb}
\usepackage{cite}
\usepackage{booktabs}
\usepackage{url}
\usepackage{fontawesome5}
\usepackage{xcolor}
\definecolor{resourcelinkblue}{RGB}{0,51,153}
\usepackage{todonotes}
\usepackage[colorlinks=true, citecolor=resourcelinkblue, linkcolor=resourcelinkblue, urlcolor=resourcelinkblue]{hyperref}
\usepackage{float}
\usepackage{comment}
\usepackage{caption}
\usepackage{subcaption}
\newcommand{\fig}[1]{\IfFileExists{#1}{\includegraphics[width=\linewidth]{#1}}{\missingfigure[figwidth=\linewidth]{#1}}}

\newcommand{\myparagraph}[1]{\noindent\textbf{#1}~}

\title{\LARGE \bf
Coding Agents for Generalized Task and Motion Planning Problems
}

\newcommand{\contactemail}{{\hypersetup{urlcolor=resourcelinkblue}\href{mailto:tsilver@princeton.edu}{\tt\small tsilver@princeton.edu}}}

\author{Matteo Merler$^{2}$, Bowen Li$^{3}$, Josh Roy$^{1}$, Yichao Liang$^{4}$, Qianwei Wang$^{1}$, Yixuan Huang$^{1}$, and Tom Silver$^{1}$%
\thanks{$^{1}$Princeton University, $^{2}$Fondazione Bruno Kessler, $^{3}$Carnegie Mellon University, $^{4}$University of Cambridge\newline
\faEnvelope[regular]\ \protect\contactemail}%
\\[0.3em]
{\small\hypersetup{urlcolor=resourcelinkblue}\faGlobe\ \href{https://agenticgentamp.github.io}{\texttt{\textbf{Website}}}\quad\textperiodcentered\quad\faCode\ \href{https://github.com/tomsilver/robocode}{\texttt{\textbf{Code}}}}}

\makeatletter
\def\bstctlcite{\@ifnextchar[{\@bstctlcite}{\@bstctlcite[@auxout]}}
\def\@bstctlcite[#1]#2{\@bsphack
  \@for\@citeb:=#2\do{%
    \edef\@citeb{\expandafter\@firstofone\@citeb}%
    \if@filesw\immediate\write\csname #1\endcsname{\string\citation{\@citeb}}\fi}%
  \@esphack}
\makeatother

\begin{document}
\bstctlcite{BSTcontrol}

\maketitle
\thispagestyle{empty}
\pagestyle{empty}

\begin{abstract}
Task and motion planning (TAMP) problems remain difficult even with full observability and object-centric states because discrete decisions are tightly coupled to geometric, kinematic, and dynamic constraints. Generalized TAMP addresses this difficulty by exploiting regularities across problem instances to reduce planning effort on new instances. However, existing methods require substantial TAMP-specific engineering. We investigate whether coding agents can automate this process by synthesizing programs that generalize across instances. Given a task description and simulator access, each agent chooses how to interact with the environment while developing a program within a fixed synthesis budget. The program is then frozen and evaluated on unseen instances. We evaluate \emph{Claude Code} (Opus 5) and \emph{Codex} (GPT-5.6 Sol and GPT-6 Astra) on 28 simulated environments from KinDER and PDDLStream, with object counts beyond those evaluated in the original benchmark. Across all program synthesis methods, we evaluate 980 generated programs on 100 held-out instances each, 98,000 evaluation episodes in total. Overall, we find that coding agents are surprisingly effective at generalized TAMP: all three agent configurations outperform hand-engineered planners, one-shot generation, and an LLM-based generalized planning baseline in mean success (56\% to 95\% versus 47\% for the planners, on the 16 environments where a planner is available). As object counts grow, the agents' programs maintain higher success than the planner, using an order of magnitude less computation per instance on average. Logs show agents using interaction to calibrate physical models, test edge cases, and refine strategies. We release all code, including the full prompts given to the agents. These findings suggest that coding agents are a strong baseline for generalized TAMP.
\end{abstract}

\section{Introduction}
We are interested in the extent to which state-of-the-art coding agents can solve the constrained manipulation problems that typify task and motion planning (TAMP). Even with full observability and object-centric states, these problems remain formally hard \cite{deshpande2019exact} and challenging in practice because horizons are long, feedback is sparse, and geometric, kinematic, and dynamic constraints are tightly coupled \cite{kaelbling2011hierarchical,srivastava2014combined,toussaint2015logic,dantam2018incremental,garrett2020pddlstream,garrett2021integrated}. To mitigate these difficulties, generalized TAMP \cite{jimenez2019review,curtis2022discovering,huang2026tamplearning,kim2018guiding,wang2021learning,wells2019learning,chitnis2016guided} exploits regularities across problem instances to produce reusable solutions, for example by learning samplers, feasibility predictors, search heuristics, or abstractions. We ask whether coding agents can similarly discover regularities that enable fast and effective planning, while relying on far less TAMP-specific scaffolding than previous methods.

\begin{figure}[!t]
	\centering
\includegraphics[width=1\columnwidth]{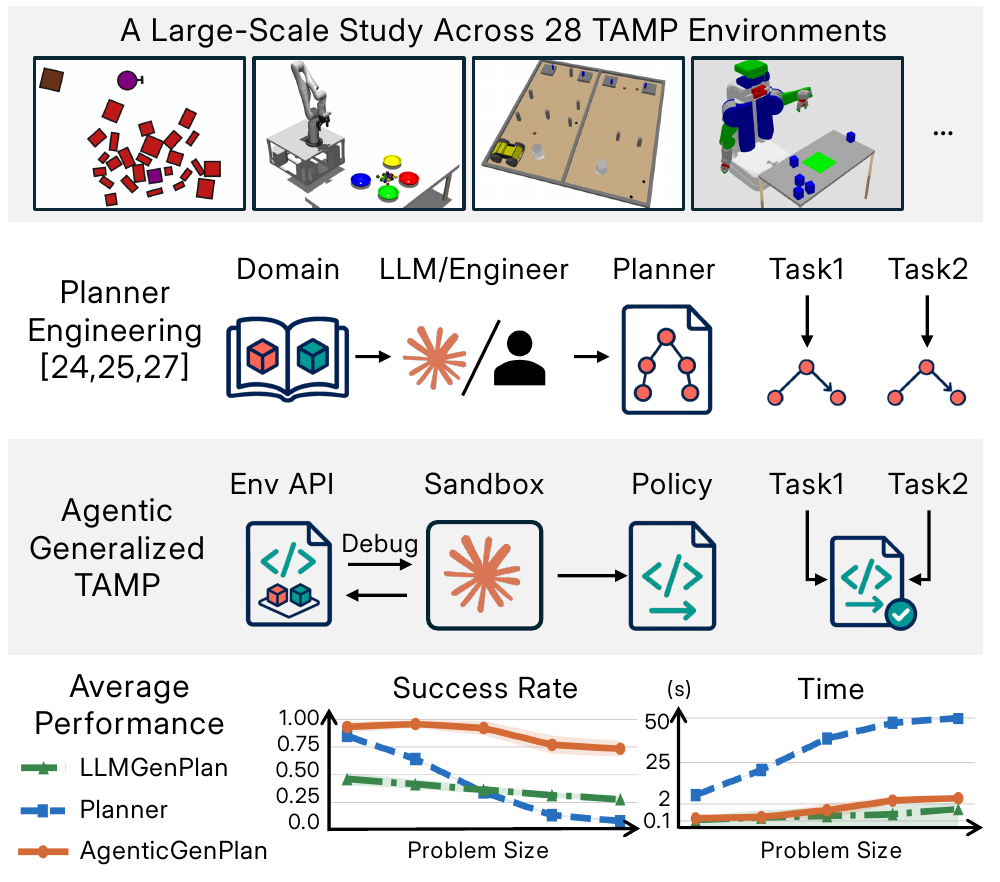}
\caption{\textbf{Coding agents for generalized TAMP.} A coding agent (\emph{Claude Code} with Opus 5) synthesizes a programmatic policy shared across problem instances, without hand-designed planning components. Compared with the planner and LLM-based generalized planner, the agentic method achieves a higher success rate while maintaining efficiency. Bottom: success rate and computation time per instance as the number of objects grows, averaged over environments with a planner available and multiple object counts.}
	\label{fig:teaser}
\end{figure}

Prior evidence points in opposite directions. Coding agents are improving rapidly on software-engineering benchmarks \cite{jimenez2024swebench,yang2024sweagent}, and large language models (LLMs) have shown increasing success in classical planning domains \cite{correa2025heuristics,correa2025frontier}. Yet when LLMs are asked to make the geometric and physical decisions within a TAMP system, they perform poorly, even when the relevant geometry is included in the prompt \cite{mendez2026systematic,huang2026kinder}. It therefore remains unclear whether advances in coding transfer to the physical reasoning that distinguishes TAMP. Either answer matters for the field. Success would suggest that coding agents can reduce much of the domain-specific engineering required by TAMP methods; failure would help identify concrete limitations of current agents.

To answer this question, we present a large-scale, systematic study of off-the-shelf coding agents for generalized TAMP (\mbox{AgenticGenPlan}). The study covers 28 environments, seven program synthesis methods, five runs per method per environment, and 100 held-out test instances per resulting program, 98,000 evaluation episodes in total. For each environment, we give a coding agent a task description, simulator access, and a fixed synthesis budget (Figure~\ref{fig:teaser}). Within this budget, the agent chooses which experiments to run, writes code to probe the simulator, and uses the results to develop and test its program. The agent returns a single program, which is frozen and evaluated on unseen instances, with no LLM involved at test time. In our main setting, the agent receives no environment source code and no hand-designed planning abstractions. Beyond the task description and simulator access, nothing is engineered for the agent.

We evaluate this approach with two different agentic backends (\emph{Claude Code} with Opus 5 and \emph{Codex} with GPT-5.6 Sol and GPT-6 Astra) on environments spanning kinematic and dynamic tasks in two and three dimensions, including the KinDER benchmark \cite{huang2026kinder} (comprising 25 environments across four families) and three PDDLStream domains \cite{mendez2026systematic}. We compare the synthesized programs with TAMP planners given the hand-designed predicates, operators, samplers, and skills supplied by the benchmarks, and we measure both success and runtime as the number of objects increases. We also compare interactive synthesis with the LLM-based generalized planning method LLMGenPlan~\cite{silver2024generalized} and the one-shot generation variant, both given environment source code. We additionally give \emph{Claude Code} and \emph{Codex} with GPT-6 Astra the environment source code. This setting serves as a reference for how the agents perform with complete knowledge of the environment. We release all code, including the full agent prompts.

Overall, all three agent configurations outperform the hand-engineered planners on the 16 environments where one is available: \emph{Claude Code} averages 82\% success, 1.7 times the planner's 47\%, and \emph{Codex} averages 95\% with GPT-6 Astra and 56\% with GPT-5.6 Sol. As object counts grow, they maintain higher success and low computation times, while the planner takes longer and solves fewer instances (Figure~\ref{fig:teaser}). All three agents also outperform one-shot generation and LLMGenPlan in mean success over all 28 environments, and source access enables successful programs in environments where main-setting runs fail.

\begin{figure*}[!tp]
    \centering
    \includegraphics[width=\textwidth]{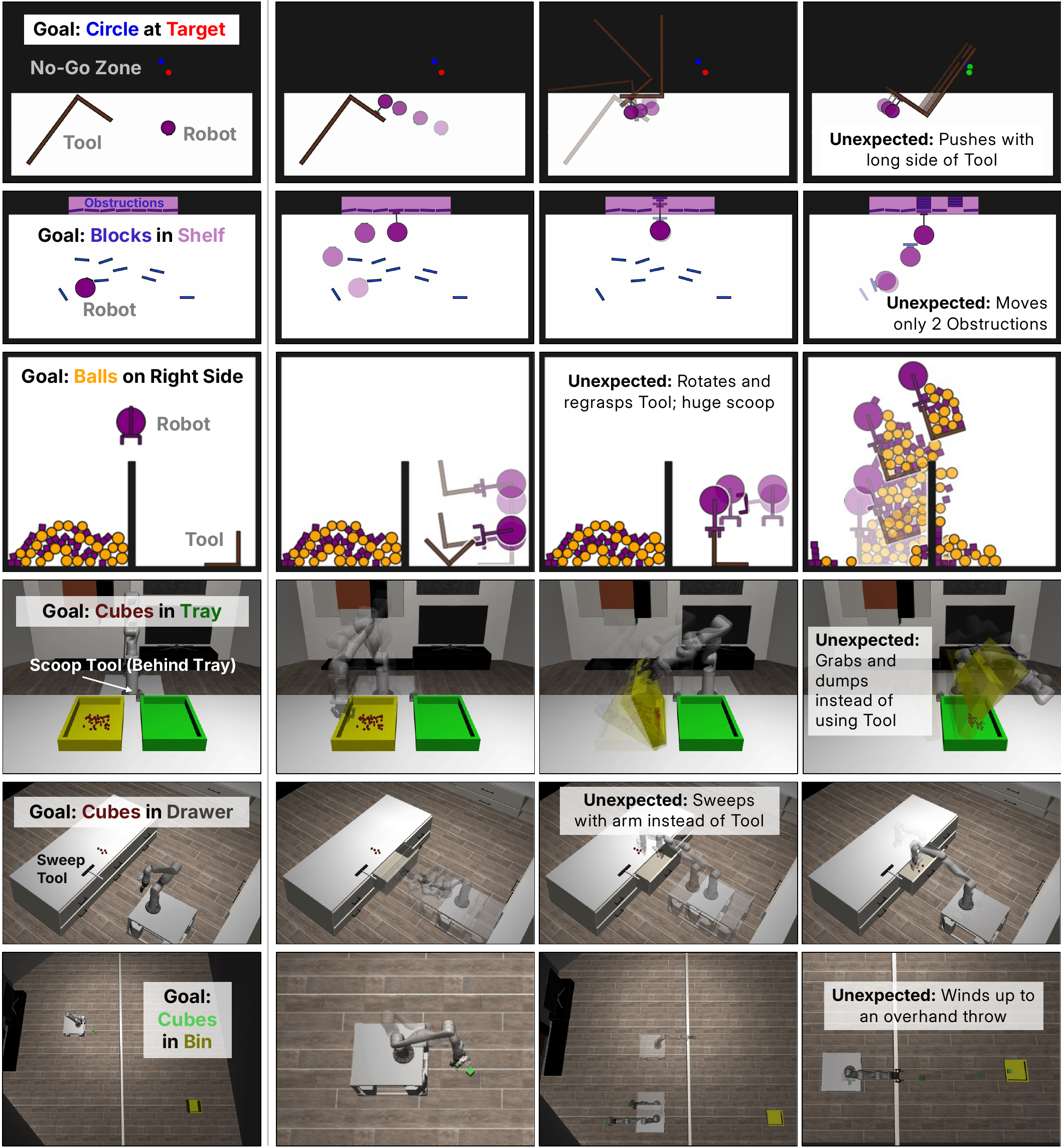}
    \caption{\textbf{Unexpected successful strategies found by the coding agents.} Each row shows one execution from left to right; the annotation describes the unexpected behavior. The strategies come from both the main setting and the + source setting.}
    \label{fig:strategies}
\end{figure*}

Analysis of the synthesized programs and interaction logs shows agents using targeted probing to infer environment dynamics and developing strategies that exploit regularities across instances. Some programs narrow a search over actions by first identifying which obstacles block a route to the goal, then planning how to move them; others adapt a fixed manipulation sequence to each instance without searching. The agents also discovered unexpected strategies, including non-prehensile maneuvers and uses of the environment layout that, to our knowledge, have not appeared in prior work on these benchmarks (Figure~\ref{fig:strategies}). We consider this qualitative evidence among the strongest in the study, both for what it says about agentic physical reasoning and because strategies absent from any published solution are hard to attribute to memorized training data. Failures remain: some dynamic three-dimensional environments are still largely unsolved in the main setting, particularly those that require sweeping or pouring many small objects. Together, these results establish coding agents as an important baseline for future work on generalized TAMP.
\section{Problem Setting}
\label{sec:prob_setting}

\subsection{MDPs and Simulator Access}

We study finite-horizon, goal-directed Markov decision processes (MDPs). An MDP is a tuple $\mathcal{M}=\langle\mathcal{S},\mathcal{A},P,R,\rho,H\rangle$, where $\mathcal{S}$ and $\mathcal{A}$ are the state and action spaces, $P$ is the transition model, $R$ is a sparse reward function indicating goal achievement, $\rho$ is the initial-state distribution, and $H$ is the horizon. States are fully observed and object-centric: a state maps each typed object to a real-valued feature vector describing properties such as pose, geometry, joint configurations, and velocity \cite{huang2026kinder}.

Each environment is represented by an MDP and a task description $d$. The initial-state distribution $\rho$ generates problem instances with varying object counts, configurations, and geometry. For example, in an object-retrieval task, instances may differ in the number and placement of obstacles surrounding the target object. The state and action spaces, transition model, and reward function are shared.

We assume simulator access through \texttt{reset} and \texttt{step}: a method can sample an initial state $s_0\sim\rho$ and execute an action $a_t$ from the current state $s_t$ to obtain $s_{t+1}\sim P(\cdot\mid s_t,a_t)$, together with the reward and termination signal.

\subsection{TAMP Environments}

We study TAMP environments that are simplified relative to real-world manipulation, following common practice in TAMP research \cite{kaelbling2011hierarchical,srivastava2014combined,toussaint2015logic,dantam2018incremental,garrett2020pddlstream,garrett2021integrated,mendez2026systematic,huang2026kinder}. With fully observed, object-centric states and no need for perception or language understanding, the remaining challenge is physical reasoning \cite{huang2026kinder}. Horizons are long, rewards are sparse, and the instance distributions are broad enough that no fixed action sequence works. Most importantly, discrete choices are tightly coupled to continuous geometric, kinematic, and dynamic constraints: which object to manipulate, which tool to use, or which subgoal to pursue determines which motions remain feasible, collision constraints grow with the number of objects, and feasible actions can occupy small regions of the action space.

\subsection{Synthesis and Evaluation}

At learning time, given the task description $d$ and simulator access, a synthesis method produces a program for $\mathcal{M}$ within a fixed budget. At step $t$, the program receives $s_t\in\mathcal{S}$ and returns $a_t\in\mathcal{A}$. Since programs can maintain internal state between steps, we denote the induced policy by $a_t=\pi(h_t)$, where $h_t=(s_0,a_0,\ldots,s_t)$ is the interaction history.

At evaluation time, the program is frozen and the synthesis method is no longer invoked. In particular, any coding agent or LLM used during learning is unavailable. The program is evaluated on initial states $s_0\sim\rho$ that were intentionally hidden during learning.

Our primary metric is success rate: the probability that the program reaches the goal within the horizon $H$ and a wall-clock limit of $\tau$ seconds. We additionally measure per-instance computation time spent during planning and deciding on actions, excluding synthesis and time spent advancing the evaluation environment.

\section{Generalized TAMP with Coding Agents}

\subsection{Coding Agents}

We instantiate the synthesis method described in Section~\ref{sec:prob_setting} using off-the-shelf coding agents from two providers: \emph{Claude Code} running Opus 5 (high), and \emph{Codex} running GPT-5.6 Sol (medium) and GPT-6 Astra (high). We choose Opus 5 as the model for all baselines, so that methods are compared with the same model. Coding agents integrate frontier LLMs with harnesses that enable them to read files, write programs, and execute arbitrary commands. We therefore run the agents inside a sandboxed Docker container with a separate filesystem and no network access. We test this isolation through red-teaming, including attempts to read environment source code, import forbidden libraries, or reach the host or network.

For our main setting, we keep the sandbox bare: the agents have only a Python interpreter with NumPy and SciPy, forcing them to generate end-to-end programs instead of relying on existing libraries. Each agent receives an initial prompt containing the task description $d$, which includes the environment name and descriptions of the task, observation and action spaces, and goal. We further explain that the environment can vary in object count, and that solutions must support any valid number of objects. We finally explain the evaluation setting, providing the agent with the time $\tau$ it will have to run its programs. We provide no hand-written TAMP predicates, operators, samplers, or skills.

To enable interactive learning, we provide the agents with simulator access to the environment: each receives a class that implements \texttt{reset} ($\rho$), \texttt{step} ($P$), and other helpers, including ones to render states as images. The agents see only a client, while the server with the actual implementation of these functions runs outside the container and is inaccessible to the agents. Using this interface, each agent chooses what to run: it can write and execute custom tests, inspect states, and revise its program based on the results.

We also evaluate an additional \emph{+ source} setting, with \emph{Claude Code} running Opus 5 and with \emph{Codex} running GPT-6 Astra, where the environment source code is available inside the container. The agents can inspect the implementation and import helper functions, \textit{e.g.}, inverse kinematics solvers. During synthesis, source access also lets the agent set arbitrary states and otherwise manipulate the simulator directly, providing generative access to the transition model \cite{kearns2002sparse}. This setting serves as a reference for how the agents perform with complete knowledge of the environment, so the agents can concentrate on developing behavior with less need to infer how the environment works through interaction.

Each agent implements the programmatic policy $\pi$ as a class with a \texttt{reset} method for episode initialization and a \texttt{get\_action} method for computing $a_t=\pi(h_t)$. Beyond this, we do not constrain the program to any particular abstractions or solution strategy, such as symbolic representations, planning algorithms, or skills. We also ask the agents to commit the program to a git repository before each test, which records its revisions so that we can later replay each one and trace how the program evolved during synthesis.

\subsection{Baselines}

We compare AgenticGenPlan against \emph{TAMP planners} that combine symbolic search with sampling to satisfy continuous constraints~\cite{garrett2021integrated}; the benchmarks provide planners for 16 of the 28 environments.
We follow official implementations~\cite{garrett2020pddlstream,huang2026kinder} for the hand-written predicates, operators, samplers, and motion planners (skills) for generating feasible trajectories. 
They compute a new plan per evaluation instance.

Furthermore, we re-implement \emph{LLMGenPlan} \cite{silver2024generalized}, which we run using Opus 5 with chain-of-thought prompting and thinking disabled, as a representative non-agentic LLM generalized planning method. In this setting, the LLM cannot use tools or access the filesystem. It receives a prompt adapted from the original work to our environment and program interfaces, together with the full source code of the environment, and must write a program implementing the same specifications as in our main setting. As the LLM cannot run code, a fixed pipeline evaluates each program generated by the LLM and returns specific pre-defined feedback (an exception with its traceback, an invalid action, or an unsolved instance with its seed). LLMGenPlan receives the same synthesis budget as the coding agents, but the LLM never chooses what to run, cannot write custom tests, and sees only the pre-defined feedback. We finally report the performance of the first program generated by LLMGenPlan as a \emph{One-shot} baseline, to evaluate the LLM without any kind of refinement loop.

\section{Experiments}

We design experiments to answer the following questions about the efficacy and efficiency of AgenticGenPlan:
\begin{enumerate}
    \item[\textbf{Q1.}] Can agents write generalized programs for TAMP?
    \item[\textbf{Q2.}] What strategies do agents discover?
    \item[\textbf{Q3.}] What advantage does being ``agentic'' give?
    \item[\textbf{Q4.}] How efficient are the synthesized programs?
    \item[\textbf{Q5.}] Does access to environment source code help?
\end{enumerate}

\begin{table*}[t]
\centering
\scriptsize
\setlength{\tabcolsep}{0.8pt}
\resizebox{\textwidth}{!}{%
\begin{tabular}{lccccccccccccccc}
\toprule
 & \multicolumn{6}{c}{\textbf{Kinematic2D}} & \multicolumn{4}{c}{\textbf{Dynamic2D}} & \multicolumn{5}{c}{\textbf{Kinematic3D}} \\
\cmidrule(lr){2-7} \cmidrule(lr){8-11} \cmidrule(lr){12-16}
\textbf{Method} & \multicolumn{1}{l}{\makebox[0pt][l]{\rotatebox{30}{\fontsize{6}{6.5}\selectfont StickButton}}} & \multicolumn{1}{l}{\makebox[0pt][l]{\rotatebox{30}{\fontsize{6}{6.5}\selectfont Obstruction}}} & \multicolumn{1}{l}{\makebox[0pt][l]{\rotatebox{30}{\fontsize{6}{6.5}\selectfont ClutteredStorage}}} & \multicolumn{1}{l}{\makebox[0pt][l]{\rotatebox{30}{\fontsize{6}{6.5}\selectfont ClutteredRetrieval}}} & \multicolumn{1}{l}{\makebox[0pt][l]{\rotatebox{30}{\fontsize{6}{6.5}\selectfont Motion}}} & \multicolumn{1}{l}{\makebox[0pt][l]{\rotatebox{30}{\fontsize{6}{6.5}\selectfont PushPullHook}}} & \multicolumn{1}{l}{\makebox[0pt][l]{\rotatebox{30}{\fontsize{6}{6.5}\selectfont Obstruction}}} & \multicolumn{1}{l}{\makebox[0pt][l]{\rotatebox{30}{\fontsize{6}{6.5}\selectfont PushPullHook}}} & \multicolumn{1}{l}{\makebox[0pt][l]{\rotatebox{30}{\fontsize{6}{6.5}\selectfont PushT}}} & \multicolumn{1}{l}{\makebox[0pt][l]{\rotatebox{30}{\fontsize{6}{6.5}\selectfont ScoopPour}}} & \multicolumn{1}{l}{\makebox[0pt][l]{\rotatebox{30}{\fontsize{6}{6.5}\selectfont Obstruction}}} & \multicolumn{1}{l}{\makebox[0pt][l]{\rotatebox{30}{\fontsize{6}{6.5}\selectfont Packing}}} & \multicolumn{1}{l}{\makebox[0pt][l]{\rotatebox{30}{\fontsize{6}{6.5}\selectfont Transport}}} & \multicolumn{1}{l}{\makebox[0pt][l]{\rotatebox{30}{\fontsize{6}{6.5}\selectfont Table}}} & \multicolumn{1}{l}{\makebox[0pt][l]{\rotatebox{30}{\fontsize{6}{6.5}\selectfont BaseMotion}}} \\
\midrule
Planner & 0.36 & 0.41 & 0.19 & 0.51 & 0.71 & -- & 0.24 & 0.13 & -- & -- & -- & 0.67 & 0.63 & -- & \textbf{1.00} \\[-2.5pt]
 & {\tiny [0.36--0.36]} & {\tiny [0.41--0.41]} & {\tiny [0.19--0.19]} & {\tiny [0.51--0.51]} & {\tiny [0.68--0.73]} & {\tiny\phantom{[0.00--0.00]}} & {\tiny [0.24--0.24]} & {\tiny [0.13--0.13]} & {\tiny\phantom{[0.00--0.00]}} & {\tiny\phantom{[0.00--0.00]}} & {\tiny\phantom{[0.00--0.00]}} & {\tiny [0.66--0.67]} & {\tiny [0.59--0.69]} & {\tiny\phantom{[0.00--0.00]}} & {\tiny [1.00--\textbf{1.00}]} \\
One-shot & 0.11 & 0.20 & 0.02 & 0.13 & 0.81 & 0.01 & 0.00 & 0.00 & 0.17 & 0.00 & 0.00 & 0.00 & 0.00 & 0.06 & 0.00 \\[-2.5pt]
{\tiny Opus 5} & {\tiny [0.03--0.16]} & {\tiny [0.00--0.54]} & {\tiny [0.00--0.06]} & {\tiny [0.02--0.23]} & {\tiny [0.17--\textbf{1.00}]} & {\tiny [0.00--0.04]} & {\tiny [0.00--0.00]} & {\tiny [0.00--0.00]} & {\tiny [0.01--0.36]} & {\tiny [0.00--0.00]} & {\tiny [0.00--0.00]} & {\tiny [0.00--0.00]} & {\tiny [0.00--0.00]} & {\tiny [0.00--0.30]} & {\tiny [0.00--0.00]} \\
LLMGenPlan & 0.83 & 0.82 & 0.02 & 0.32 & 0.97 & 0.12 & 0.49 & 0.00 & 0.17 & 0.00 & 0.00 & 0.14 & 0.05 & 0.85 & \textbf{1.00} \\[-2.5pt]
{\tiny Opus 5} & {\tiny [0.70--0.97]} & {\tiny [0.53--\textbf{1.00}]} & {\tiny [0.00--0.06]} & {\tiny [0.21--0.41]} & {\tiny [0.95--\textbf{1.00}]} & {\tiny [0.02--0.34]} & {\tiny [0.36--0.77]} & {\tiny [0.00--0.00]} & {\tiny [0.00--0.41]} & {\tiny [0.00--0.00]} & {\tiny [0.00--0.00]} & {\tiny [0.00--0.58]} & {\tiny [0.00--0.27]} & {\tiny [0.41--\textbf{1.00}]} & {\tiny [1.00--\textbf{1.00}]} \\
AgenticGenPlan & 0.94 & 0.96 & 0.37 & 0.21 & \textbf{1.00} & 0.69 & 0.77 & 0.52 & 0.76 & 0.83 & 0.26 & 0.37 & 0.17 & 0.99 & \textbf{1.00} \\[-2.5pt]
{\tiny GPT-5.6 Sol (medium)} & {\tiny [0.90--0.99]} & {\tiny [0.93--0.99]} & {\tiny [0.19--0.65]} & {\tiny [0.12--0.27]} & {\tiny [1.00--\textbf{1.00}]} & {\tiny [0.46--0.90]} & {\tiny [0.60--0.90]} & {\tiny [0.37--0.64]} & {\tiny [0.21--\textbf{1.00}]} & {\tiny [0.20--\textbf{1.00}]} & {\tiny [0.00--0.69]} & {\tiny [0.00--0.65]} & {\tiny [0.00--0.53]} & {\tiny [0.97--\textbf{1.00}]} & {\tiny [1.00--\textbf{1.00}]} \\
AgenticGenPlan & \textbf{1.00} & \textbf{1.00} & 0.99 & 0.81 & \textbf{1.00} & 0.98 & 0.88 & 0.83 & \textbf{1.00} & 0.97 & 0.85 & 0.66 & \textbf{0.99} & \textbf{1.00} & \textbf{1.00} \\[-2.5pt]
{\tiny Opus 5 (high)} & {\tiny [1.00--\textbf{1.00}]} & {\tiny [1.00--\textbf{1.00}]} & {\tiny [0.99--\textbf{1.00}]} & {\tiny [0.36--0.98]} & {\tiny [1.00--\textbf{1.00}]} & {\tiny [0.91--\textbf{1.00}]} & {\tiny [0.79--0.94]} & {\tiny [0.63--0.95]} & {\tiny [1.00--\textbf{1.00}]} & {\tiny [0.89--\textbf{1.00}]} & {\tiny [0.41--\textbf{1.00}]} & {\tiny [0.52--0.79]} & {\tiny [0.94--\textbf{1.00}]} & {\tiny [1.00--\textbf{1.00}]} & {\tiny [1.00--\textbf{1.00}]} \\
AgenticGenPlan & \textbf{1.00} & 1.00 & \textbf{1.00} & \textbf{0.96} & \textbf{1.00} & \textbf{1.00} & \textbf{0.96} & \textbf{0.91} & 1.00 & \textbf{1.00} & \textbf{0.97} & \textbf{0.89} & 0.78 & \textbf{1.00} & \textbf{1.00} \\[-2.5pt]
{\tiny GPT-6 Astra (high)} & {\tiny [1.00--\textbf{1.00}]} & {\tiny [0.99--\textbf{1.00}]} & {\tiny [0.99--\textbf{1.00}]} & {\tiny [0.91--\textbf{1.00}]} & {\tiny [1.00--\textbf{1.00}]} & {\tiny [1.00--\textbf{1.00}]} & {\tiny [0.94--\textbf{0.98}]} & {\tiny [0.79--\textbf{1.00}]} & {\tiny [0.99--\textbf{1.00}]} & {\tiny [0.99--\textbf{1.00}]} & {\tiny [0.92--0.99]} & {\tiny [0.67--\textbf{1.00}]} & {\tiny [0.00--\textbf{1.00}]} & {\tiny [1.00--\textbf{1.00}]} & {\tiny [1.00--\textbf{1.00}]} \\
\midrule
AgenticGenPlan + source & 1.00 & 1.00 & 1.00 & 0.96 & 1.00 & 1.00 & 0.92 & 0.92 & 1.00 & 0.88 & 1.00 & 0.98 & 1.00 & 1.00 & 1.00 \\[-2.5pt]
{\tiny Opus 5 (high)} & {\tiny [1.00--1.00]} & {\tiny [1.00--1.00]} & {\tiny [1.00--1.00]} & {\tiny [0.92--0.99]} & {\tiny [1.00--1.00]} & {\tiny [1.00--1.00]} & {\tiny [0.87--0.96]} & {\tiny [0.77--0.99]} & {\tiny [1.00--1.00]} & {\tiny [0.65--1.00]} & {\tiny [0.99--1.00]} & {\tiny [0.94--1.00]} & {\tiny [1.00--1.00]} & {\tiny [1.00--1.00]} & {\tiny [1.00--1.00]} \\
AgenticGenPlan + source & 1.00 & 1.00 & 1.00 & 1.00 & 1.00 & 1.00 & 0.93 & 0.97 & 1.00 & 0.97 & 1.00 & 1.00 & 1.00 & 1.00 & 1.00 \\[-2.5pt]
{\tiny GPT-6 Astra (high)} & {\tiny [1.00--1.00]} & {\tiny [1.00--1.00]} & {\tiny [1.00--1.00]} & {\tiny [1.00--1.00]} & {\tiny [1.00--1.00]} & {\tiny [1.00--1.00]} & {\tiny [0.83--0.98]} & {\tiny [0.91--1.00]} & {\tiny [1.00--1.00]} & {\tiny [0.87--1.00]} & {\tiny [1.00--1.00]} & {\tiny [1.00--1.00]} & {\tiny [1.00--1.00]} & {\tiny [1.00--1.00]} & {\tiny [1.00--1.00]} \\
\bottomrule
\end{tabular}
}
\caption{\textbf{Success rate over environments}. We report the mean over five runs, with [min--max] across run-level success rates below. Bold marks the best mean and the best max per environment among the main-setting rows. AgenticGenPlan + source additionally gives the agent the environment source code. A dash marks environments for which KinDER provides no planner. Dynamic3D and PDDLStream environments are listed in Table~\ref{tab:main-2}.}
\label{tab:main}
\end{table*}

\begin{table*}[t]
\centering
\scriptsize
\setlength{\tabcolsep}{0.8pt}
\resizebox{\textwidth}{!}{%
\begin{tabular}{lccccccccccccc}
\toprule
 & \multicolumn{10}{c}{\textbf{Dynamic3D}} & \multicolumn{3}{c}{\textbf{PDDLStream}} \\
\cmidrule(lr){2-11} \cmidrule(lr){12-14}
\textbf{Method} & \multicolumn{1}{l}{\makebox[0pt][l]{\rotatebox{30}{\fontsize{6}{6.5}\selectfont BalanceBeam}}} & \multicolumn{1}{l}{\makebox[0pt][l]{\rotatebox{30}{\fontsize{6}{6.5}\selectfont ConstrainedCupboard}}} & \multicolumn{1}{l}{\makebox[0pt][l]{\rotatebox{30}{\fontsize{6}{6.5}\selectfont Dynamo}}} & \multicolumn{1}{l}{\makebox[0pt][l]{\rotatebox{30}{\fontsize{6}{6.5}\selectfont Rearrange}}} & \multicolumn{1}{l}{\makebox[0pt][l]{\rotatebox{30}{\fontsize{6}{6.5}\selectfont ScoopPour}}} & \multicolumn{1}{l}{\makebox[0pt][l]{\rotatebox{30}{\fontsize{6}{6.5}\selectfont Shelf}}} & \multicolumn{1}{l}{\makebox[0pt][l]{\rotatebox{30}{\fontsize{6}{6.5}\selectfont SortClutteredBlocks}}} & \multicolumn{1}{l}{\makebox[0pt][l]{\rotatebox{30}{\fontsize{6}{6.5}\selectfont SweepIntoDrawer}}} & \multicolumn{1}{l}{\makebox[0pt][l]{\rotatebox{30}{\fontsize{6}{6.5}\selectfont SweepSimple}}} & \multicolumn{1}{l}{\makebox[0pt][l]{\rotatebox{30}{\fontsize{6}{6.5}\selectfont Tossing}}} & \multicolumn{1}{l}{\makebox[0pt][l]{\rotatebox{30}{\fontsize{6}{6.5}\selectfont Packing}}} & \multicolumn{1}{l}{\makebox[0pt][l]{\rotatebox{30}{\fontsize{6}{6.5}\selectfont Blocked}}} & \multicolumn{1}{l}{\makebox[0pt][l]{\rotatebox{30}{\fontsize{6}{6.5}\selectfont Rovers}}} \\
\midrule
Planner & -- & -- & -- & -- & -- & 0.27 & -- & 0.00 & -- & 0.77 & 0.54 & 0.74 & 0.30 \\[-2.5pt]
 & {\tiny\phantom{[0.00--0.00]}} & {\tiny\phantom{[0.00--0.00]}} & {\tiny\phantom{[0.00--0.00]}} & {\tiny\phantom{[0.00--0.00]}} & {\tiny\phantom{[0.00--0.00]}} & {\tiny [0.22--0.30]} & {\tiny\phantom{[0.00--0.00]}} & {\tiny [0.00--0.00]} & {\tiny\phantom{[0.00--0.00]}} & {\tiny [0.77--0.78]} & {\tiny [0.50--0.56]} & {\tiny [0.70--0.77]} & {\tiny [0.24--0.38]} \\
One-shot & 0.00 & 0.00 & 0.13 & 0.00 & 0.00 & 0.00 & 0.00 & 0.00 & 0.00 & 0.00 & 0.12 & 0.00 & 0.26 \\[-2.5pt]
{\tiny Opus 5} & {\tiny [0.00--0.00]} & {\tiny [0.00--0.00]} & {\tiny [0.01--0.34]} & {\tiny [0.00--0.00]} & {\tiny [0.00--0.00]} & {\tiny [0.00--0.00]} & {\tiny [0.00--0.00]} & {\tiny [0.00--0.00]} & {\tiny [0.00--0.00]} & {\tiny [0.00--0.00]} & {\tiny [0.00--0.40]} & {\tiny [0.00--0.01]} & {\tiny [0.00--0.87]} \\
LLMGenPlan & 0.00 & 0.00 & 0.98 & 0.00 & 0.00 & 0.00 & 0.00 & 0.00 & 0.00 & 0.00 & 0.21 & 0.00 & 0.86 \\[-2.5pt]
{\tiny Opus 5} & {\tiny [0.00--0.00]} & {\tiny [0.00--0.00]} & {\tiny [0.95--\textbf{1.00}]} & {\tiny [0.00--0.00]} & {\tiny [0.00--0.00]} & {\tiny [0.00--0.00]} & {\tiny [0.00--0.00]} & {\tiny [0.00--0.00]} & {\tiny [0.00--0.00]} & {\tiny [0.00--0.01]} & {\tiny [0.00--0.39]} & {\tiny [0.00--0.01]} & {\tiny [0.81--0.97]} \\
AgenticGenPlan & 0.64 & 0.00 & \textbf{1.00} & 0.09 & 0.00 & 0.24 & 0.01 & 0.00 & 0.00 & 0.00 & 0.81 & 0.75 & 0.78 \\[-2.5pt]
{\tiny GPT-5.6 Sol (medium)} & {\tiny [0.00--\textbf{1.00}]} & {\tiny [0.00--0.00]} & {\tiny [1.00--\textbf{1.00}]} & {\tiny [0.00--0.32]} & {\tiny [0.00--0.00]} & {\tiny [0.00--0.63]} & {\tiny [0.00--0.03]} & {\tiny [0.00--0.00]} & {\tiny [0.00--0.00]} & {\tiny [0.00--0.00]} & {\tiny [0.41--\textbf{1.00}]} & {\tiny [0.34--0.99]} & {\tiny [0.00--\textbf{1.00}]} \\
AgenticGenPlan & 0.97 & 0.08 & 0.92 & 0.58 & 0.09 & 0.60 & 0.30 & 0.00 & 0.00 & \textbf{0.99} & 0.99 & 0.38 & 0.98 \\[-2.5pt]
{\tiny Opus 5 (high)} & {\tiny [0.87--\textbf{1.00}]} & {\tiny [0.00--0.34]} & {\tiny [0.60--\textbf{1.00}]} & {\tiny [0.01--0.96]} & {\tiny [0.00--\textbf{0.31}]} & {\tiny [0.00--\textbf{1.00}]} & {\tiny [0.13--\textbf{0.72}]} & {\tiny [0.00--0.00]} & {\tiny [0.00--0.00]} & {\tiny [0.96--\textbf{1.00}]} & {\tiny [0.96--\textbf{1.00}]} & {\tiny [0.25--0.56]} & {\tiny [0.90--\textbf{1.00}]} \\
AgenticGenPlan & \textbf{1.00} & \textbf{0.29} & \textbf{1.00} & \textbf{0.96} & \textbf{0.14} & \textbf{1.00} & \textbf{0.43} & \textbf{0.72} & \textbf{0.03} & 0.96 & \textbf{1.00} & \textbf{0.99} & \textbf{1.00} \\[-2.5pt]
{\tiny GPT-6 Astra (high)} & {\tiny [0.99--\textbf{1.00}]} & {\tiny [0.04--\textbf{0.63}]} & {\tiny [1.00--\textbf{1.00}]} & {\tiny [0.93--\textbf{0.98}]} & {\tiny [0.00--0.23]} & {\tiny [0.98--\textbf{1.00}]} & {\tiny [0.31--0.50]} & {\tiny [0.15--\textbf{1.00}]} & {\tiny [0.00--\textbf{0.15}]} & {\tiny [0.82--\textbf{1.00}]} & {\tiny [1.00--\textbf{1.00}]} & {\tiny [0.98--\textbf{1.00}]} & {\tiny [0.99--\textbf{1.00}]} \\
\midrule
AgenticGenPlan + source & 1.00 & 0.47 & 1.00 & 0.77 & 0.38 & 0.43 & 0.49 & 0.57 & 0.07 & 0.85 & 1.00 & 0.87 & 0.99 \\[-2.5pt]
{\tiny Opus 5 (high)} & {\tiny [0.99--1.00]} & {\tiny [0.05--1.00]} & {\tiny [1.00--1.00]} & {\tiny [0.15--0.97]} & {\tiny [0.21--0.57]} & {\tiny [0.00--0.98]} & {\tiny [0.20--0.96]} & {\tiny [0.00--0.97]} & {\tiny [0.00--0.21]} & {\tiny [0.31--1.00]} & {\tiny [1.00--1.00]} & {\tiny [0.43--1.00]} & {\tiny [0.98--1.00]} \\
AgenticGenPlan + source & 1.00 & 1.00 & 1.00 & 0.99 & 0.42 & 1.00 & 0.95 & 0.86 & 0.44 & 0.98 & 1.00 & 1.00 & 1.00 \\[-2.5pt]
{\tiny GPT-6 Astra (high)} & {\tiny [0.99--1.00]} & {\tiny [0.99--1.00]} & {\tiny [1.00--1.00]} & {\tiny [0.97--1.00]} & {\tiny [0.06--1.00]} & {\tiny [0.99--1.00]} & {\tiny [0.89--0.99]} & {\tiny [0.47--1.00]} & {\tiny [0.35--0.50]} & {\tiny [0.93--1.00]} & {\tiny [0.98--1.00]} & {\tiny [1.00--1.00]} & {\tiny [1.00--1.00]} \\
\bottomrule
\end{tabular}
}
\caption{\textbf{Dynamic3D and PDDLStream environments}. Same protocol and notation as Table~\ref{tab:main}. The planner for the PDDLStream environments is PDDLStream itself.}
\label{tab:main-2}
\end{table*}

\subsection{Environment Setup}

We select the KinDER benchmark \cite{huang2026kinder} as our primary benchmark. This covers 25 environments, grouped into four families: \emph{Kinematic2D}, \emph{Dynamic2D}, \emph{Kinematic3D}, and \emph{Dynamic3D}.
The kinematic environments are free of dynamics, while in the dynamic environments, outcomes depend on contact, velocity, and friction, so successful programs may need behaviors such as sweeping, pouring, and tossing.
Nineteen of these environments feature variants with different object counts, including counts beyond those evaluated in the original benchmark.
We further include three PDDLStream domains, originally introduced by Garrett et al. \cite{garrett2020pddlstream}, where LLMs have been shown ineffective \cite{mendez2026systematic}: Packing, Blocked, and Rovers.

For each method, we perform five independent runs per environment. Each coding-agent and LLMGenPlan synthesis run has a budget of \$20 in model usage. We evaluate the resulting programs and planners on the same 100 instances per environment, sampled from the same initial-state distribution $\rho$ used during synthesis. We generate the evaluation seeds randomly to make it unlikely that agents test them during synthesis, and verify afterward that none were used. We use a timeout $\tau$ of 60 seconds per evaluation instance for all methods, matching the original KinDER protocol.

\begin{figure*}[!tp]
    \centering
    \includegraphics[width=\textwidth,trim=0 158.27bp 0 0,clip]{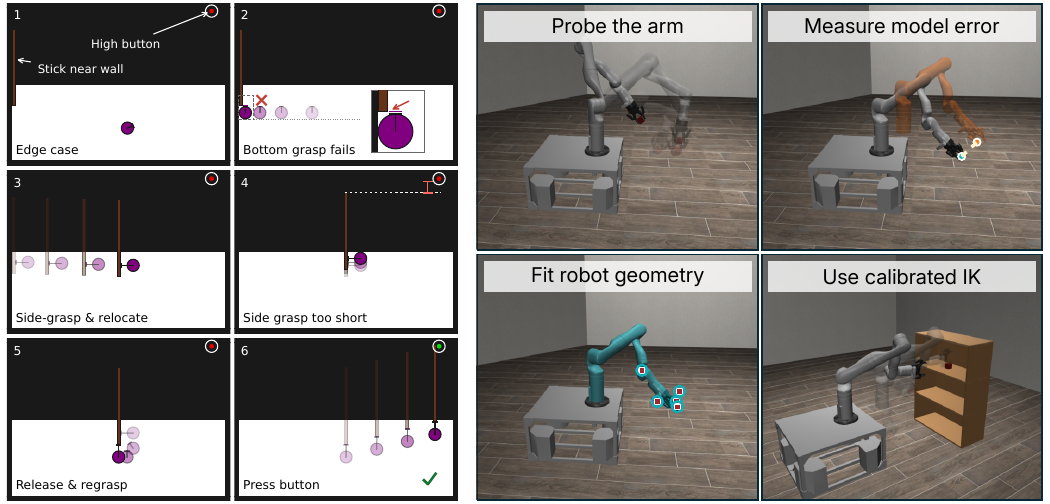}
    \caption{\textbf{Left: custom tests reveal edge cases in StickButton.} \emph{Opus} writes a script that repeatedly calls \texttt{reset} with different seeds to find wall-adjacent sticks and high buttons for testing its program. The wall prevents a bottom grasp. A side grasp lets the robot move the stick away from the wall but leaves the high button out of reach; releasing and re-grasping from below provides the required height. \textbf{Right: learning robot geometry through interaction.} In Shelf, \emph{Opus} probes the arm while holding a cube, fits its kinematic model to observed cube positions, and uses the calibrated model for inverse kinematics (IK). The initial model discrepancy is exaggerated for visibility.}
    \label{fig:stickbutton-edge-cases}
    \label{fig:shelf-discovery}
\end{figure*}

\subsection{Results and Analysis}
Tables~\ref{tab:main} and~\ref{tab:main-2} report success rates across the 28 environments. Below, we write \emph{Opus}, \emph{Sol}, and \emph{Astra} for AgenticGenPlan with \emph{Claude Code} running Opus 5 and \emph{Codex} running GPT-5.6 Sol and GPT-6 Astra, respectively, and \emph{Opus} + source and \emph{Astra} + source for AgenticGenPlan + source. \emph{Astra} is the best performing agent, averaging 99\% success on Kinematic2D, 97\% on Dynamic2D, 93\% on Kinematic3D, 65\% on the harder Dynamic3D, and nearly 100\% on PDDLStream. \emph{Opus} follows, with 96\%, 92\%, 90\%, 45\%, and 78\%, respectively. All three agent configurations exceed the planning baseline in most environments with one available (15 of 16 for \emph{Astra}, 12 for \emph{Opus}, and 9 for \emph{Sol}), without the hand-designed models, skills, and samplers that the planners use, and their programs exploit regularities within each environment to restrict the decisions considered at test time (\textbf{Q1}).

\myparagraph{Discovering Unexpected Strategies:}
We find that the coding agents discover unexpected manipulation strategies, depicted in Figure~\ref{fig:strategies} (\textbf{Q2}).
To name a few, in Dynamic2D ScoopPour (row three), an \emph{Opus} program in the main setting rotates the tool and regrasps it from the left side, which lets it scoop far more balls at once.
In SweepIntoDrawer (row five), an \emph{Opus} + source program ignores the tool (sweeper) and uses the gripper to sweep the cubes one by one.

\myparagraph{Testing Edge Cases:}
One advantage that agentic methods provide over static LLM calls is the flexibility to write and execute custom tests (\textbf{Q3}).
These let agents investigate failures and evaluate their policies on selected configurations.
In StickButton, for example, \emph{Opus} writes a script that repeatedly calls \texttt{reset} with different seeds, inspects the sampled states, and saves edge cases involving wall-adjacent sticks and high buttons. It combines these edge cases with typical instances to build a diverse test suite, then tests its programs on both to challenge them across a broad range of configurations (Figure~\ref{fig:stickbutton-edge-cases}, left).

\myparagraph{Building Internal Models:}
Compared to static LLM-based synthesis methods, we also find that AgenticGenPlan draws on prior robotics knowledge to build initial models, then tests and calibrates them through interaction (\textbf{Q3}).
In Shelf, for example, one \emph{Opus} run constructs an initial kinematic model of the Kinova Gen3 arm from the agent's own knowledge, without internet access.
It then uses a grasped cube as a marker because the state exposes object positions but not the hand's position. Moving the arm through different configurations lets it calibrate the model's predictions of cube positions from joint angles (Figure~\ref{fig:shelf-discovery}, right). From a few observations, it fits six parameters describing the robot mount and grasp offsets, reducing the RMSE between predicted and observed cube positions from 38.9 to 1.8~mm on the calibration observations. It later uses the fitted model to solve inverse kinematics (IK) as part of the solution.

Building on what they learn about the environment, agents continue interacting with it to refine control parameters and manipulation strategies. In BalanceBeam, for example, one \emph{Sol} run moves its small-block placement targets closer to the beam's center, increasing success from 6\% to 64\%.

\begin{figure*}[!tp]
    \centering
    \includegraphics[width=\textwidth]{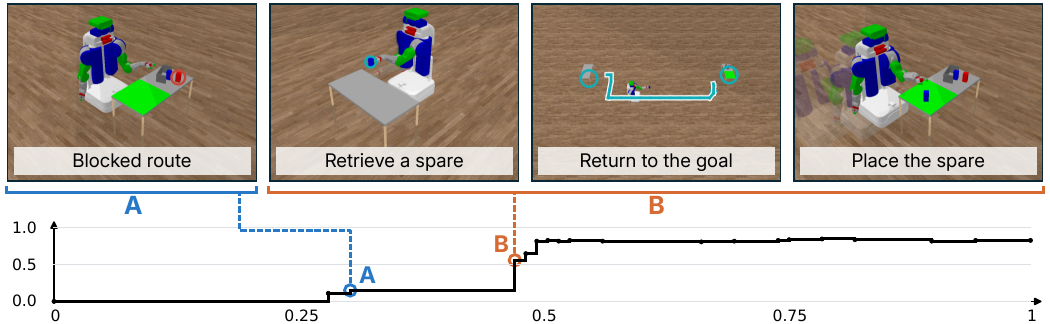}
    \caption{\textbf{Refining manipulation strategies during synthesis.} In one \emph{Sol} Blocked run, a revision adding a spare-block fallback changes success from 15\% (A) to 56\% (B); the pictured execution retrieves an alternative goal block from a distant table. The curve shows the held-out success rate of each commit, in commit order.}
    \label{fig:policy-refinement}
\end{figure*}

\myparagraph{Comparing Agents:}
\emph{Astra} achieves the highest mean success, above \emph{Opus} in 20 of 28 environments. In 11 of these 20, the best \emph{Opus} program scores at least as high as the best \emph{Astra} program, so the gain comes from greater consistency: in Shelf, for example, \emph{Opus} programs range from 0\% to 100\% success. The weakest never discovers which shelf the cubes must go on, and another places one cube on the correct shelf but leaves the rest on the floor in front of the shelf. In contrast, every \emph{Astra} program is tested and revised on instances with up to eight cubes and scores at least 98\%. In other environments, \emph{Astra} finds strategies that no \emph{Opus} program uses. In SweepIntoDrawer, where no \emph{Opus} program succeeds, the best \emph{Astra} program solves every instance by picking up the cubes instead of sweeping them, while the only \emph{Astra} program that sweeps scores 15\%. In Kinematic3D Packing, instances with three parts combine a cube with two triangles, each either right or equilateral, on a small rack. The best \emph{Opus} program places right triangles in fixed, hand-tuned positions but has no way to place equilateral ones. Three \emph{Astra} programs instead search over positions and rotations for every part and solve all instances. Some \emph{Opus} programs are still better: in Dynamic3D ScoopPour, all agents skip scooping and move the whole tray at once, and the best \emph{Opus} program tips the tray over while resting it on the target tray, whereas the best \emph{Astra} program flips the tray entirely in mid-air and cannot recover when it drops it.

Between the other two agents, \emph{Opus} is stronger than \emph{Sol}, achieving higher mean success in 22 of 28 environments, with four ties. \emph{Sol} is higher only in Dynamo and in Blocked, where it reaches 75\%, compared with 38\% for \emph{Opus} and 74\% for the planner. In Blocked, the robot can retrieve either a nearby obstructed block or, when available, an unobstructed block on a distant table. \emph{Opus} programs often persist with the nearby block, where obstacle orientations make grasping and retrieval difficult. \emph{Sol} programs can instead switch to the distant block after unsuccessful attempts. In one \emph{Sol} run, a commit adding this fallback raises success from 15\% to 56\%, with 41 newly solved instances and no lost successes (Figure~\ref{fig:policy-refinement}). Subsequent commits reach 83\% final success, solving all 80 instances with an alternative block but only three of the 20 without one. \emph{Astra} programs combine both behaviors: one run grasps the nearby block reliably, even at awkward orientations, and falls back to the distant table when a grasp fails, solving all 100 instances.

\begin{table}[t]
\centering
\small
\setlength{\tabcolsep}{3pt}
\begin{tabular}{lc}
\toprule
\textbf{Method} &
\textbf{Time/action (ms)} \\
\midrule
AgenticGenPlan & 11.741 \\[-2.5pt]
{\tiny Opus 5 (high)} & {\tiny [0.041--109.403]} \\
AgenticGenPlan + source & 42.480 \\[-2.5pt]
{\tiny Opus 5 (high)} & {\tiny [0.056--381.509]} \\
\midrule
AgenticGenPlan & 1.299 \\[-2.5pt]
{\tiny GPT-6 Astra (high)} & {\tiny [0.033--6.047]} \\
AgenticGenPlan + source & 50.392 \\[-2.5pt]
{\tiny GPT-6 Astra (high)} & {\tiny [0.053--477.045]} \\
\bottomrule
\end{tabular}
\caption{\textbf{Policy-computation time per action} on the 15 environments
where \emph{Opus}, \emph{Opus} + source, \emph{Astra}, and \emph{Astra} +
source each reach 100\% held-out success in at least one run. Each method is
averaged over its own perfect runs within each environment; we report the
mean across environments, with [min--max] across environment-level means.
Source access yields slower programs on average, as many of them invoke the
environment source code as part of planning at decision time.}
\label{tab:efficiency}
\end{table}

\myparagraph{Program Computation Efficiency:}
Table~\ref{tab:efficiency} compares policy-computation time per action on the 15 environments where \emph{Opus}, \emph{Opus} + source, \emph{Astra}, and \emph{Astra} + source each reach 100\% held-out success in at least one run, averaging each method over its own perfect runs (\textbf{Q4}). Main-setting \emph{Opus} programs choose actions quickly, averaging 11.7~ms per action. \emph{Opus} + source programs average over 3 times that, with large variation across environments, from over twice as fast to nearly 200 times slower, because many of them invoke the environment source code as part of planning. Synthesis without source code instead forces the agent to distill what it learns into self-contained code, which yields faster programs at evaluation time. \emph{Astra} programs are the fastest, averaging 1.3~ms per action, about 9 times faster than \emph{Opus}, and source access costs them even more, raising the average to 50.4~ms. \emph{Astra} programs tend to use compact, closed-form control, such as a few hardcoded joint poses or analytic inverse kinematics, whereas \emph{Opus} programs more often search over large precomputed candidate sets, with retry and fallback logic, at decision time. Programs also need far less computation than the planners: on the 14 environments with a planner available and multiple object counts, \emph{Opus} and \emph{Astra} programs take 2.1~s and 0.5~s per instance on average, compared with 29~s for the planners.

\myparagraph{Environment Source Code Access:}
We further inspect whether providing the agents with environment source code helps them generate programs (\textbf{Q5}). Source access raises mean success for both agents, from 74\% to 84\% for \emph{Opus} and from 86\% to 95\% for \emph{Astra}. Source access is not sufficient on its own: LLMGenPlan also receives the source code, uses Opus 5, and has the same \$20 budget, yet reaches 28\%, while \emph{Opus} + source is higher in 27 of 28 environments. Most of this gap points to the benefit of agentic synthesis, where the agent can run code, test, and revise (\textbf{Q3}). Environment source code provides both information and implementations that programs can reuse. With source access, the agent can read goals and success checks from the code, whereas in the main setting the agent must infer them from rewards and rendered images, and sometimes settles on an inaccurate goal and writes its program against it. In SortClutteredBlocks, for example, main-setting \emph{Astra} programs assign cubes to bins by their order in the state, with a rule that works for four cubes but fails on every twenty-cube instance. With source access, programs read each cube's target bin (43\% to 95\%). Programs also reuse the implementation directly, for example loading the robot's model for inverse kinematics, calling motion planners and collision checks, or testing actions in private simulators. In Kinematic3D Packing, one \emph{Opus} run uses internal collision information to identify that the gripper, rather than the held part, collides with the rack. It then changes the grasp to provide clearance. In SweepIntoDrawer, where \emph{Opus} and \emph{Sol} score zero in the main setting, source access raises \emph{Opus}'s mean success to 57\%, with one run reaching 97\%.

\section{Related Work}
\label{sec:related-work}

\subsection{Generalized Planning and Learning for TAMP}

TAMP couples discrete decisions with continuous feasibility constraints~\cite{garrett2021integrated}. Systems such as PDDLStream provide interfaces between symbolic planning and procedures for sampling and checking continuous quantities~\cite{garrett2020pddlstream}. Generalized planning seeks solutions that apply across related problem instances~\cite{jimenez2019review}. In TAMP, this objective has motivated learning reusable samplers, feasibility predictors, search heuristics, and state and action abstractions~\cite{kim2018guiding,wells2019learning,chitnis2016guided,curtis2022discovering}; see~\cite{huang2026tamplearning} for a recent survey. We ask whether general-purpose coding agents can likewise exploit regularities across instances without being confined by specific TAMP abstractions.

\subsection{Foundation Models for Task and Motion Planning}
One branch of foundation model-driven TAMP research leverages large language models (LLMs) to guide a TAMP planner.
For example, Text2Motion combines language-guided task planning with skill affordances~\cite{lin2023text2motion}, and LLM$^3$ uses motion-planning failures to guide plan revision~\cite{wang2024llm3}. 
In addition to guidance, LLM-based program synthesis can also automate the engineering of TAMP components. 
PRoC3S generates programs whose continuous parameters are resolved through constraint satisfaction~\cite{curtis2025proc3s}. 
MOPS searches over programs specifying constraints for trajectory optimization~\cite{shcherba2025mops}, while OWL-TAMP uses vision-language models to generate constraints within a TAMP system~\cite{kumar2026owltamp}. 
Instead of committing to certain TAMP abstractions, our evaluation leaves the internal structure of the generated solution open: an agent may implement search, optimization, sampling, or task-specific procedures on its own, with the final objective of efficiently solving as many problem instances as possible.

\subsection{Agentic Synthesis of Robotic Programs}

Since Code-as-Policies~\cite{liang2023codeaspolicies}, LLMs have demonstrated the ability to generate robot control code, integrating manipulation skills and perception interfaces. Later works such as GenCHiP~\cite{burns2024genchip} and InstructFlow~\cite{chi2025instructflow} motivate code as a flexible policy representation, while highlighting the importance of the supplied action interface and feedback. In symbolic task planning, Silver et al.~\cite{silver2024generalized} use LLMs and execution feedback to synthesize reusable programs that solve new PDDL instances without further LLM calls.

Recent agentic systems provide even closer precedents for reusable policy synthesis. RHO introduces a reflective evolutionary optimizer over policy repositories, using coding agents and execution feedback~\cite{elmaaroufi2026rho}. ASPIRE develops mechanisms for discovering, repairing, and reusing robotic skills~\cite{lu2026aspire}, and MEMENTO introduces memory-guided evolutionary search over policy programs~\cite{sygkounas2026memento}. 
The benchmarks used in these works, such as Robosuite~\cite{zhu2020robosuite} and LIBERO~\cite{liu2023libero}, focus on general-purpose manipulation rather than the physical reasoning challenges of TAMP problems~\cite{huang2026kinder}.
Our focus is a systematic assessment of off-the-shelf coding agents generating code as generalized policies for constrained TAMP-like environments.

\subsection{Systematic Evaluation of Embodied Agents}
Mendez-Mendez's study of LLMs within PDDLStream is the closest precedent in evaluation focus~\cite{mendez2026systematic}. It examines configurations in which LLMs replace task planning, continuous sampling, or both within established planning architectures, and compares them with engineered planners. 
Whereas it queries LLMs during problem solving, we use simulator feedback to develop a program, then freeze it for LLM-free evaluation on unseen instances.
Several benchmarks and systematic studies also inform our evaluation. 
CaP-X benchmarks coding agents for general manipulation control~\cite{fu2026capx}.
Tsui et al.~\cite{tsui2026faea} evaluate an LLM agent SDK on general manipulation tasks with iterative execution.
KinDER~\cite{huang2026kinder} provides procedurally generated environments designed to isolate physical reasoning and already compares planning and learning approaches; we build on these environments and their baseline implementations.

\section{Discussion}

\myparagraph{Limitations:}
Our setup assumes fully observed, object-centric states and simulator access during synthesis. It therefore does not establish performance under perception uncertainty.
The models' training data are undisclosed, so prior exposure to benchmark code cannot be excluded. 
However, the synthesis logs show policies being developed through environment probing, testing, and substantial revision. Together with the unexpected strategies the agents synthesized, this provides evidence against simply recalling complete solutions seen during pretraining.

\myparagraph{Conclusions:}
We show that coding agents are strong generalized TAMP planners, synthesizing reusable programs that outperform hand-engineered planners and existing LLM-based synthesis methods on our benchmarks. These results extend the promise of LLM-based generalized planning to environments with geometric, kinematic, and dynamic constraints. Through interaction, agents can investigate the physical behavior of an environment and develop effective strategies without being supplied with its symbolic models, skills, or samplers. This ability to discover both how an environment works and how to solve its tasks makes coding agents an important baseline for future TAMP research.

\section*{Acknowledgments}

We thank Pietro Ferrazzi for feedback on a draft of this paper. This work was partially supported by a Princeton SEAS Innovation Grant, an NVIDIA Academic Grant Program, and a Princeton AI Lab Grant.

\bibliographystyle{IEEEtran}
\bibliography{references}

\end{document}